\documentclass[letterpaper, 10 pt, conference]{ieeeconf}  

\IEEEoverridecommandlockouts                              

\usepackage{graphics} 
\usepackage{amsmath} 
\usepackage{amssymb}  
\usepackage{bm}
\usepackage{eso-pic}
\usepackage[most]{tcolorbox}
\usepackage{graphicx}

\usepackage{algorithm}
\usepackage{algorithmic}
\usepackage{multirow}

\usepackage{makeidx}
\usepackage{cite}
\usepackage{url}
\usepackage{here}
\usepackage{color, hyperref}
\hypersetup{
    colorlinks=true,
    citecolor=blue,
    linkcolor=black,
    urlcolor=blue,
}
\usepackage{umoline}

\title{\LARGE \bf
Real-World Robot Control by Deep Active Inference\\ 
with a Temporally Hierarchical World Model
}

\author{Kentaro Fujii$^{1}$ and Shingo Murata$^{1}$
\thanks{*This work was supported by JST PRESTO (JPMJPR22C9), {JSPS KAKENHI} (JP24K03012), Mori Manufacturing Research and Technology Foundation.}
\thanks{$^{1}$Kentaro Fujii and Shingo Murata are with Graduate School of Integrated Design Engineering, Keio University {\tt\small oakwood.n14.4sp@keio.jp, murata@elec.keio.ac.jp}}%
}

\IEEEoverridecommandlockouts

\newcommand{\IEEEcopyrightfooter}{%
  \footnotesize
  \textcopyright~2025~IEEE. Permission from IEEE must be obtained for all other uses, in any current or future media, including reprinting/republishing this material for advertising or promotional purposes, creating new collective works, for resale or redistribution to servers or lists, or reuse of any copyrighted component of this work in other works.\\
  The final version of this paper is available at: https://doi.org/10.1109/LRA.2025.3636032
}

\newtcolorbox{copyrightbox}{%
  enhanced,
  sharp corners,        
  colback=white,
  colframe=black,          
  boxrule=0.4pt,              
  arc=2mm,                    
  left=1.5mm,right=1.5mm,top=1.5mm,bottom=1.5mm, 
  boxsep=0mm,
}

\AddToShipoutPictureFG*{%
  \AtPageLowerLeft{%
    \raisebox{10mm}{
      \makebox[\paperwidth]{%
        \begin{minipage}{0.96\paperwidth}
          \begin{copyrightbox}
            \centering \IEEEcopyrightfooter
          \end{copyrightbox}
        \end{minipage}%
      }%
    }%
  }%
}
\begin{document}

\maketitle
\thispagestyle{empty}
\pagestyle{empty}

\begin{abstract}

Robots in uncertain real‑world environments must perform both goal‑directed and exploratory actions. However, most deep learning-based control methods neglect exploration and struggle under uncertainty. To address this, we adopt deep active inference, a framework that accounts for human goal-directed and exploratory actions. Yet, conventional deep active inference approaches face challenges due to limited environmental representation capacity and high computational cost in action selection. We propose a novel deep active inference framework that consists of a world model, an action model, and an abstract world model. The world model encodes environmental dynamics into hidden state representations at slow and fast timescales. The action model compresses action sequences into abstract actions using vector quantization, and the abstract world model predicts future slow states conditioned on the abstract action, enabling low-cost action selection. We evaluate the framework on object-manipulation tasks with a real-world robot. 
Results show that it achieves high success rates across diverse manipulation tasks and switches between goal-directed and exploratory actions in uncertain settings, while making action selection computationally tractable.
These findings highlight the importance of modeling multiple timescale dynamics and abstracting actions and state transitions.

\end{abstract}

\section{INTRODUCTION}

With recent advances in deep learning-based robot control methods, there is growing expectation for the realization of robots capable of achieving a wide range of human-like goals \cite{chi2023diffusion, etukuru2024robot, black2024pi}.
In real-world environments, the presence or arrangement of objects required for a task is often uncertain, and current robots struggle to cope with such uncertainty \cite{lynch2023interactive}.
In contrast, humans can not only act toward achieving goals but also explore to resolve environmental uncertainty—e.g., by searching for the location of an object—thereby adapting effectively to uncertain situations \cite{friston2015epistemic, Friston2017}.

To realize robots capable of both goal-directed and exploratory actions, we focus on deep active inference \cite{Millidge2020, fountas2020deep, Mazzaglia2021, fujii2024real}—a deep learning-based framework grounded in a computational theory that accounts for various cognitive functions \cite{friston2010free, friston2015epistemic, schwartenbeck2019computational}.
However, deep active inference faces two key challenges: (1) its performance heavily depends on the capability of the framework to represent environmental dynamics \cite{sajid2021exploration}, and (2) the computational cost is prohibitively high \cite{Mazzaglia2021}, making it difficult to apply to real-world robots.

To address these challenges, we propose a deep active inference framework comprising a world model, an action model, and an abstract world model.
The world model learns hidden state transitions to represent environmental dynamics from human-collected robot action and observation data \cite{ha2018world, Hafner2018, ahmadi2019novel}.
The action model maps a sequence of actual actions to one of a learned set of abstract actions, each corresponding to a meaningful behavior (e.g., moving an object from a dish to a pan) \cite{lee2024behavior}.
The abstract world model learns the relationship between the state representations learned by the world model and the abstract action representations learned by the action model \cite{gumbsch2024learning}.
By leveraging the abstract world model and the abstract action representations, the framework enables efficient active inference.

To evaluate the proposed method, we conducted robot experiments in real-world environments with uncertainty.
We investigated whether the framework could reduce computational cost, enable the robot to achieve diverse goals involving the manipulation of multiple objects, and perform exploratory actions to resolve environmental uncertainty.

\begin{figure*}
    \centering
    \includegraphics[width=0.90\linewidth]{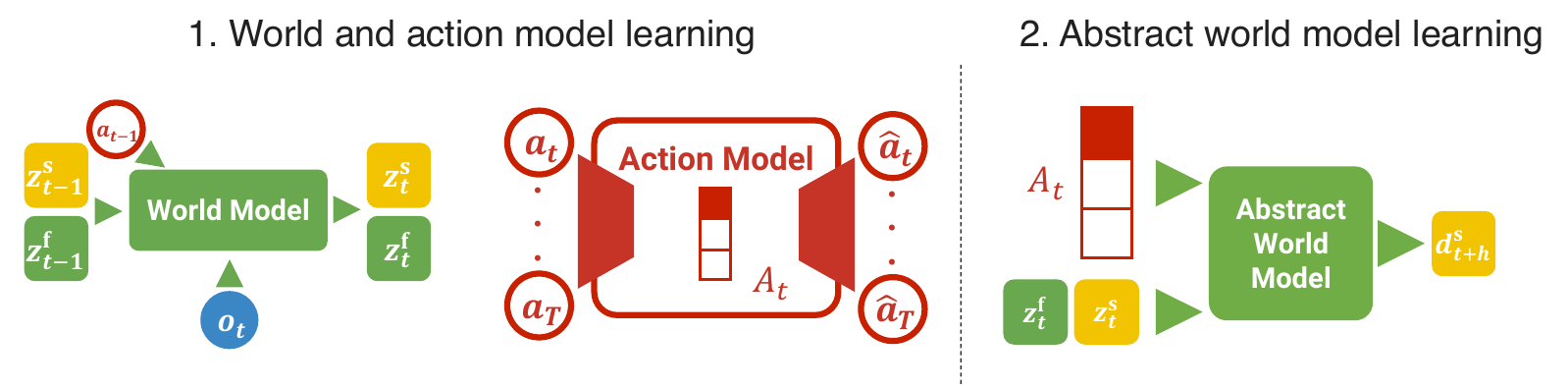}
    \caption{The overview of the proposed framework. The framework comprises a world model, an action model, and an abstract world model. Here, key variables are visualized: observation $o_t$ and action $a_t$ are processed by the world model to infer hierarchical hidden states $z_t^\mathrm{s}$, $z_t^\mathrm{f}$. The action model compresses action sequences into abstract actions $A_t$. The abstract world model uses $A_t$ to predict the future slow deterministic state $d_{t+h}^\mathrm{s}$.
    }
    \label{fig:framework}
\end{figure*}

\section{RELATED WORK}
\subsection{Learning from Demonstration (LfD) for Robot Control}
LfD is a method to train robots by imitating human experts, providing safe, task-relevant data for learning control policies \cite{ravichandar2020recent, correia2024survey, zare2024survey, florence2022implicit, lancaster2023modem, Jang2022}.
A key advancement contributing to recent progress in LfD for robotics is the idea of generating multi-step action sequences, rather than only single-step actions \cite{zhao2023learning, chi2023diffusion, lee2024behavior, etukuru2024robot, black2024pi}.
However, a major challenge in LfD is the difficulty of generalizing to environments with uncertainty, even when trained on large amounts of expert demonstrations \cite{lynch2023interactive}.
In this work, we focus on the approach that uses quantized features extracted from action sequences \cite{lee2024behavior}, and treat the extracted features as abstract action representations.

\subsection{World Model}
A world model captures the dynamics of the environment by modeling the relationship between data (observations), their latent causes (hidden states), and actions.
They have recently attracted significant attention in the context of model-based reinforcement learning \cite{ha2018world, Hafner2018}, especially in artificial agents and robotics \cite{Taniguchi03072023}.
However, when robots learn using a world model, their performance is constrained by the model’s capability to represent environmental dynamics \cite{Cai2022, deng2024facing}.
In particular, learning long-term dependencies in the environment remains a challenge.
One solution is to introduce temporal hierarchy into the model structure \cite{Kim2019, Cai2022, Saxena2021, fujii2023hierarchical}.
Furthermore, by incorporating abstract action representations that capture slow dynamics, the model can more efficiently predict future observations and states \cite{gumbsch2024learning}.
Temporal hierarchy can be introduced by differentiating state update frequencies \cite{Kim2019, Saxena2021, Cai2022} or modulating time constants of state transitions \cite{Yamashita2008, ahmadi2019novel, spieler2024the}. 
In this work, we adopt the latter to better represent slow dynamics in our world model \cite{fujii2023hierarchical}.

\section{THE FORMULATION OF ACTIVE INFERENCE}
The free-energy principle \cite{friston2010free, friston2015epistemic, Friston2017} is a computational principle that accounts for various cognitive functions.
According to this principle, human observations $o$ are generated by unobservable hidden states $z$, which evolve in response to actions $a$, following a partially observable Markov decision process \cite{friston2015epistemic}.
The brain is assumed to model this generative process with the world model.
Under the free-energy principle, human perception and action aim to minimize the surprise $-\log p(o)$. 
However, since directly minimizing surprise is intractable, active inference instead minimizes its tractable upper bound, the variational free energy \cite{friston2015epistemic, Friston2017}.

Perception can be formulated as the minimization of the following variational free energy at time step $t$ \cite{Mazzaglia2021, mazzaglia2022free, Smith2022}:
\begin{equation}
    \label{eq:vfe}
    \begin{aligned}
        \mathcal{F}(t)&=D_{\mathrm{KL}}\left[q\left(z_t \right)\| p\left(z_t\right)\right]-\mathbb{E}_{q\left(z_t\right)}\left[\log p\left(o_t \mid z_t\right)\right]\\
        &\geq -\log p(o_t).
    \end{aligned}
\end{equation}
Here, 
$q(z_t)$ denotes the approximate posterior over the hidden state $z_t$, 
$D_\mathrm{KL}[q(\cdot)||p(\cdot)]$ is the Kullback–Leibler (KL) divergence. Note that the first line of \eqref{eq:vfe} is equivalent to the negative evidence lower bound \cite{Kingma2013, rezende2014stochastic}.

Action can be formulated as the minimization of expected free energy (EFE), which extends variational free energy to account for future states and observations.
Let $\tau > t$ be a future time step, The EFE is defined as follows \cite{Smith2022}:
\begin{equation}
    \label{eq:efe}
    \begin{aligned}
        \mathcal{G}(\tau)\approx&-\underbrace{\mathbb{E}_{q(o_\tau, z_\tau \mid \pi)}[\log q(z_\tau \mid o_\tau, \pi)-\log q(\mathrm{z_\tau} \mid \pi)]}_{\text{Epistemic value}}\\
        &-\underbrace{\mathbb{E}_{q(o_\tau \mid \pi)}[\log p(o_\tau \mid o_{\text{pref}})]}_{\text{Extrinsic value}}.
    \end{aligned}
\end{equation}
Here, the expectation is over the observation $o_\tau$ because the future observation is not yet available \cite{Smith2022}, and $\pi$ indicates the policy (i.e. an action sequence).
The variable $o_{\text{pref}}$ is referred to as a preference, which encodes the goal, and the distribution $p(o_\tau \mid o_{\text{pref}})$ is called the prior preference.
In \eqref{eq:efe}, the first term referred to as the epistemic value is the mutual information between the state $z_\tau$ and the observation $o_\tau$.
This term encourages exploratory policies that reduce the uncertainty in the prior belief $q(z_\tau \mid \pi)$.
On the other hand, the second term referred to as the extrinsic value encourages goal-directed policies.
Therefore, selecting a policy $\pi$ that minimizes the EFE can account for both exploratory and goal-directed actions \cite{friston2015epistemic, Friston2017, igari2024selection}.

Conventional active inference requires calculating the EFE over all possible action sequences during task execution, which is intractable for real-world action spaces \cite{Friston2017}. 
Recent works have addressed this by using the EFE as a loss function for training of action generation models \cite{Millidge2020, fountas2020deep, Mazzaglia2021}, but often ignored exploration capability.
In this work, we propose a novel framework focusing on both goal-achievement performance and exploration capability tractably calculating the EFE during task execution.
\label{sec:formulation}

\begin{figure}[t]
    \centering
    \includegraphics[width=0.85\linewidth]{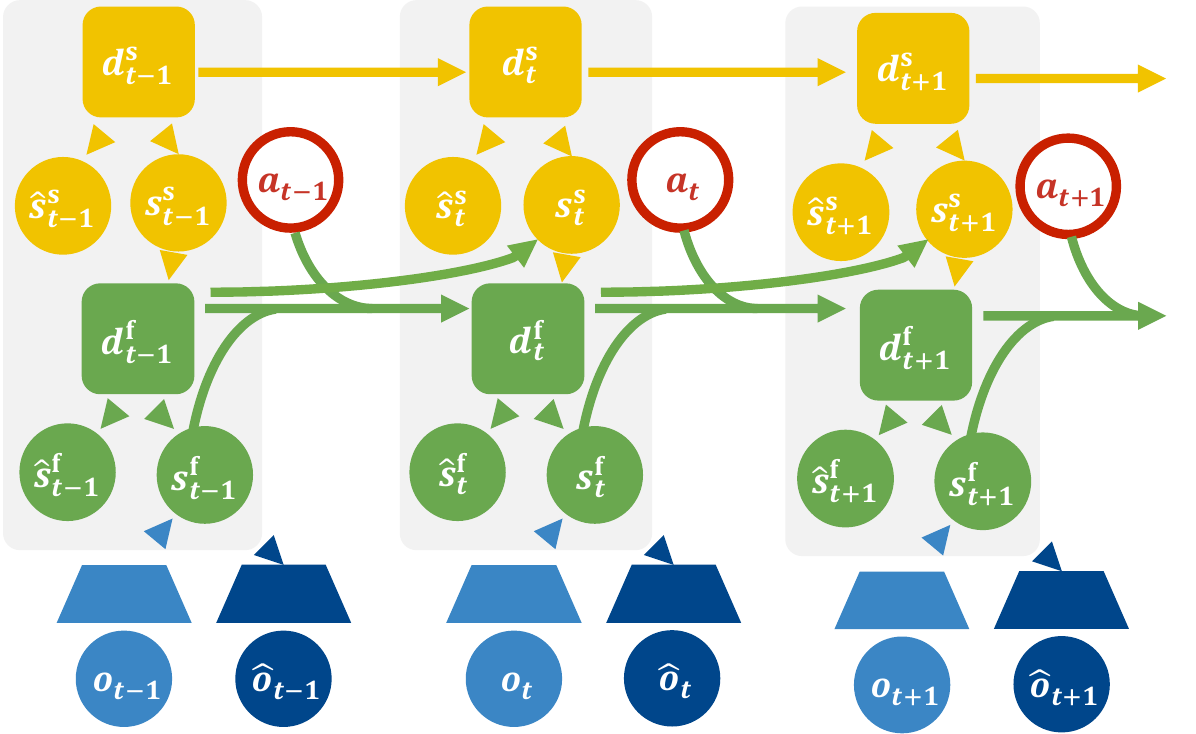}
    \caption{The world model. It consists of a dynamics model, an encoder, and a decoder. The dynamics model has two different timescales. }
    \label{fig:world}
\end{figure}
\section{METHOD}

\subsection{Framework}

We propose a framework based on deep active inference that enables both goal achievement and exploration. The proposed framework consists of a world model, an action model, and an abstract world model (Fig. \ref{fig:framework}). 

\subsubsection{World Model}

The world model comprises a dynamics model, an encoder, and a decoder, all of which are trained simultaneously (Fig. \ref{fig:world}). 
As the dynamics model, we utilize a hierarchical model \cite{fujii2022hierarchical}, which consists of the slow and fast states as the hidden states $z_t = \{z^\mathrm{s}_t, z^\mathrm{f}_t\}$ for time step $t$. Both deterministic $d$ and stochastic $s$ states are defined for each of the slow and fast states $z^\mathrm{s}_t=\{d^\mathrm{s}_t, s^\mathrm{s}_t\}, z^\mathrm{f}_t=\{d^\mathrm{f}_t, s^\mathrm{f}_t\}$, respectively.
These hidden states are calculated as follows:
\begin{equation}
    \begin{aligned} 
        &\text{Slow dynamics}\\
        &\quad \text{Deterministic state:} &&\quad d^\mathrm{s}_t=f^\mathrm{s}_\theta\left(z^\mathrm{s}_{t-1}\right)\\
        &\quad \text{Prior:} &&\quad \hat{s}^\mathrm{s}_t \sim p^\mathrm{s}_\theta\left(s^\mathrm{s}_t \mid d^\mathrm{s}_t\right)\\
        &\quad \text{Approximate posterior:} &&\quad s^\mathrm{s}_t \sim q^\mathrm{s}_\theta\left(s^\mathrm{s}_t \mid d^\mathrm{s}_t, d^\mathrm{f}_{t-1}\right).\\
        &\text{Fast dynamics}\\
        &\quad \text{Deterministic State:} &&\quad d^\mathrm{f}_t=f^\mathrm{f}_\theta\left(s^\mathrm{s}_t, z^\mathrm{f}_{t-1}, a_{t-1}\right)\\
        &\quad \text{Prior:} &&\quad \hat{s}^\mathrm{f}_t \sim p^\mathrm{f}_\theta\left(s^\mathrm{f}_t \mid d^\mathrm{f}_t\right)\\
        &\quad \text{Approximate posterior:} &&\quad s^\mathrm{f}_t \sim q^\mathrm{f}_\theta\left(s^\mathrm{f}_t \mid d^\mathrm{f}_t, o_t\right)\\
    \end{aligned}
\end{equation}
Here, $o_t$ is the observation and $a_{t-1}$ is the action at the previous time step.
The approximate posterior of the fast dynamics $q^\mathrm{f}_\theta$ is conditioned on the observation $o_t$ by receiving its features extracted by the encoder.

The slow and fast deterministic states $d^\mathrm{s}_t$ and $d^\mathrm{f}_t$ are computed by multiple timescale recurrent neural network parameterized with a time constant \cite{Yamashita2008}.
When the time constant is large, the state tends to evolve slowly compared to when the time constant is small.
Therefore, by setting the time constant for the slow layer larger than one for the fast layer, the dynamics model represent a temporal hierarchy.
The slow and fast stochastic states $s^\mathrm{s}_t, \hat{s}^\mathrm{s}_t$ and $s^\mathrm{f}_t, \hat{s}^\mathrm{f}_t$ are represented as one-hot vectors sampled from an approximate posterior or a prior, 
defined by categorical distributions \cite{Hafner2020}.

The decoder is employed to reconstruct the observation $o_t$ from the hidden state $z_t$, modeling likelihood $p_\theta(o_t \mid z_t)$. 
Simultaneously, a network $p_\theta(d^\mathrm{f}_t\mid z_t^\mathrm{s})$ that predicts the fast deterministic state $\hat{d}^\mathrm{f}_t$ from the slow hidden state $z^\mathrm{s}_t$ is also trained. 
The predicted deterministic state $\hat{d}^\mathrm{f}_t$ is then used to sample the fast stochastic state. By combining both slow and predicted fast hidden states as inputs to the decoder, the dynamics model can represent the observation likelihood $p_\theta(o_t \mid z^\mathrm{s}_t)$ \footnote{Correctly, this distribution is written as $p_\theta\left(o_t \mid z_t^{\mathrm{s}}\right)=\int p_\theta\left(o_t \mid z_t\right) p_\theta^{\mathrm{f}}\left(s_t^{\mathrm{f}} \mid d_t^{\mathrm{f}}\right) p_\theta\left(d_t^{\mathrm{f}} \mid z_t^{\mathrm{s}}\right) \mathrm{d} z_t^{\mathrm{f}}$. We approximate the marginal over the fast states $z_t^{\mathrm{f}}$ with a single Monte Carlo sample.} 
based on only the slow hidden state $z^\mathrm{s}_t$.

The world model is trained by minimizing the variational free energy $\mathcal{F}(t)$. Here, since the fast deterministic state $d^\mathrm{f}_t$ can be regarded as an observation for the slow dynamics, the variational free energies $\mathcal{F}_\mathrm{s}(t)$ and $\mathcal{F}_\mathrm{f}(t)$ can be computed separately for the slow and fast layers, respectively. 
Furthermore, we also minimize, as an auxiliary task, the negative log-likelihood of observation $o_t$ given the slow hidden state $z^\mathrm{s}_t$, denoted as $\log p_\theta(o_t \mid z^\mathrm{s}_t)$.
In summary, the variational free energy $\mathcal{F}(t)$ in this work is described as follows:
\begin{equation}
    \begin{aligned}
        \mathcal{F}(t)=&\mathcal{F}_\mathrm{s}(t)+\mathcal{F}_\mathrm{f}(t)-\log p_\theta(o_t|z^\mathrm{s}_t)\\
        \mathcal{F}_\mathrm{s}(t)=&\quad  D_\text{KL}[\mathrm{sg}(q^\mathrm{s}_\theta\left(s^\mathrm{s}_t \mid d^\mathrm{s}_t, d^\mathrm{f}_{t-1}\right)) \|p^\mathrm{s}_\theta\left(s^\mathrm{s}_t \mid d^\mathrm{s}_t\right)]\\
        &-\log p_\theta(d^\mathrm{f}_t\mid z_t^\mathrm{s})\\
        \mathcal{F}_\mathrm{f}(t)=&\quad D_\text{KL}[\mathrm{sg}(q^\mathrm{f}_\theta\left(s^\mathrm{f}_t \mid d^\mathrm{f}_t, o_t\right)) \|p^\mathrm{f}_\theta\left(s^\mathrm{f}_t \mid d^\mathrm{f}_t\right)]\\
        &-\log p_\theta(o_t\mid z_t)].
    \end{aligned}
\end{equation}
Here, for the KL divergence calculation, we use the KL balancing technique with a weighting factor $w$ \cite{Hafner2020}.

\subsubsection{Action Model}

The action model consists of an encoder $\mathcal{E}_\phi$ and a decoder $\mathcal{D}_\phi$ composed of multilayer perceptron (MLP), as well as a residual vector quantizer \cite{van2017neural,zeghidour2021soundstream, lee2024behavior} $\mathcal{Q}_\phi$ with $N_q=2$ layers. 
First, the encoder $\mathcal{E}_\phi$ embeds the action sequence $a_{t:t+h}$ of length $h$ into a low-dimensional feature $A_t$. Next, the feature $A_t$ is quantized into $\hat{A}_t$ using the residual vector quantizer $\mathcal{Q}_\phi$. The residual vector quantizer includes codebooks $\{C_i\}_{i=1}^{N_q}$, each containing $K$ learnable codes $\{c_{i,j}\}_{j=1}^K$. 
Specifically, the quantized vector at layer $i$ is the code $c_{i,k}$ having the smallest Euclidean distance to the input at layer $i$.
The quantized feature $\hat{A}_t$ is the sum of outputs from each quantization layer $\{\hat{A}_{t,i}\}_{i=1}^{N_q}=\sum_i^{N_q}c_{i,k}$. 
Finally, the decoder $\mathcal{D}_\phi$ reconstructs the quantized feature $\hat{A}_t$ into the action sequence $\hat{a}_{t:t+h}$. In summary, the procedure of the action model is described as follows:
\begin{equation}
    \begin{aligned}
        A_t &= \mathcal{E}_\phi(a_{t:t+h})\\
        \hat{A}_t &= \mathcal{Q}_\phi(A_t)\\
        \hat{a}_{t:t+h}&=\mathcal{D}_\phi(\hat{A}_t).
    \end{aligned}
\end{equation}
We treat the feature $\hat{A}_t$, obtained by the action model, as an abstract action representing the action sequence $a_{t:t+h}$.

The encoder $\mathcal{E}_\phi$ and decoder $\mathcal{D}_\phi$ of the action model are trained by minimizing the following objective:
\begin{equation}
    \begin{aligned}
        \mathcal{L}_\phi=\quad&\lambda_\text{MSE} \|a_{t:t+h}-\hat{a}_{t:t+h}\|_2^2\\+&\lambda_{\text{commit}}\Sigma_{i=1}^{N_q}\left\|(A_{t}-\Sigma_i(\hat{A}_{t,i-1}))-\operatorname{sg}\left(c_{i,k}\right)\right\|_2^2\\
    \end{aligned}
\end{equation}
where we assume $\hat{A}_{t,0}=0$. Moreover, $\lambda_\text{MSE}$ and $\lambda_{\text{commit}}$ are coefficients for the reconstruction loss $\mathcal{L}_\text{MSE}$ and the commitment loss $\mathcal{L}_\text{commit}$, respectively. The learning of the codebooks $\{C_i\}_{i=1}^{N_q}$ of  the residual vector quantizer $\mathcal{Q}_\phi$ is performed using exponential moving averages \cite{van2017neural, lee2024behavior}.

\subsubsection{Abstract World Model}
{The abstract world model $\mathcal{W}_\psi$ learns a mapping from the current world model state $z_t$ and an abstract action $A_t$ to the future slow deterministic state $d^{\mathrm{s}}_{t+h}$.
In other words, it provides an abstract representation of state transitions.}
The model $\mathcal{W}_\psi$ is composed of MLP and takes the abstract action $A_t$ and the current world model state $z_t$ as inputs to predict the slow deterministic state $d^\mathrm{s}_{t+h}$.
Here, the input abstract action $A_t$ to $\mathcal{W}_\psi$ can be any of the $K^{N_q}$ combinations of learned codes from the action model, denoted as $\{\hat{A}_{n}\}_{n=1}^{K^{N_q}}$.
Accordingly, for a given current hidden state $z_t$, the abstract world model $\mathcal{W}_\psi$ predicts $K^{N_q}$ possible future slow deterministic states $\{d^\mathrm{s}_{t+h,n}\}_{n=1}^{K^{N_q}}$:
\begin{equation}
    \begin{aligned}
        \{{\hat{d}^\mathrm{s}_{t+h,n}}\}_{n=1}^{K^{N_q}} = \mathcal{W}_\psi(z_t, \{\hat{A}_{n}\}_{n=1}^{K^{N_q}}).
    \end{aligned}
\end{equation}
The abstract world mode is trained by minimizing the following objective:
\begin{equation}
    \mathcal{L}_{\psi}=\frac{1}{K^{N_q}}\sum_{n=1}^{K^{N_q}}\|{\hat{d}^\mathrm{s}_{t+h,n}}-d^\mathrm{s}_{t+h,n}\|_2^2.
\end{equation}
Here, to obtain the target slow deterministic states $\{d^\mathrm{s}_{t+h,n}\}_{n=1}^{K^{N_q}}$, we utilize latent imagination of the world model \cite{Hafner2018}. To this end, the action sequences $\{\hat{a}_{0:h,n}\}_{n=1}^{K^{N_q}}$ are generated from the code combinations $\{\hat{A}_{n}\}_{n=1}^{K^{N_q}}$ using the decoder $\mathcal{D}_\phi$ of the action model. Then, by leveraging the prior distribution over the fast states, the slow deterministic states $\{d^\mathrm{s}_{t+h,n}\}_{n=1}^{K^{N_q}}$ at $h$ steps ahead are obtained. 

\begin{figure*}[t]
    \centering
    \includegraphics[width=0.85\linewidth]{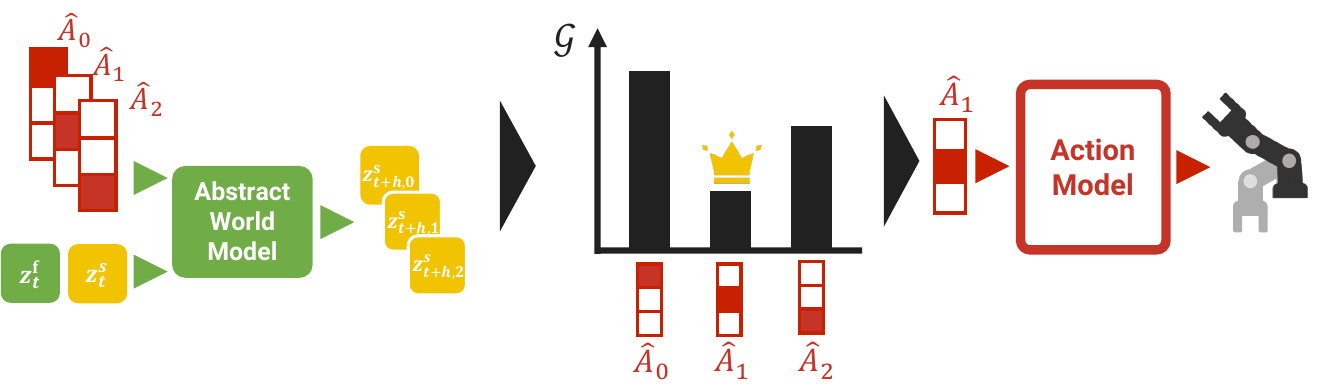}
    \caption{Action selection based on the minimization of EFE.
             First, future states are predicted for multiple abstract actions.
             Then, the EFE is calculated for each of the predicted future states.
             Finally, the robot execute action sequence reconstructed from the abstract action that yields the lowest EFE.}
    \label{fig:efe-calc}
\end{figure*}
\subsection{Action Selection}
To make the EFE $\mathcal{G}(\tau)$ calculation tractable, our framework leverages a learned, finite set of abstract actions $\{\hat{A}_{n}\}_{n=1}^{K^{N_q}}$, instead of considering all possible (and thus infinite) continuous action sequences.

First, we reformulate \eqref{eq:efe} in accordance with our world model (for a detailed derivation, see Appendix \ref{appendix:efe}):
\begin{equation}
    \label{eq:actual-efe}
    \begin{aligned}
        \mathcal{G}(\tau)=&-\mathbb{E}_{q_\theta(o_\tau, z_\tau \mid \pi)}[\log q_\theta(z_\tau \mid o_\tau, \pi)-\log q_\theta(\mathrm{z_\tau} \mid \pi)]\\
        &-\mathbb{E}_{q_\theta(o_\tau \mid \pi)}[\log p(o_\tau \mid  o_{\text{pref}})]\\
        \approx&-\mathbb{E}_{q_\theta(o_\tau, z_\tau \mid \pi)}[\log q_\theta(s^{\mathrm{f}}_\tau \mid z_\tau^\mathrm{s}, o_\tau )-\log q_\theta(s^{\mathrm{f}}_\tau \mid z_\tau^\mathrm{s})]\\
        &-\mathbb{E}_{q_\theta(o_\tau, z_\tau \mid \pi)}[\log p(o_\tau \mid  o_{\text{pref}})].
    \end{aligned}
\end{equation}
Here, the joint distribution $q_\theta(o_\tau, z_\tau \mid \pi)$ can be decomposed as
$q_\theta(o_\tau, z_\tau \mid \pi) = p_\theta(o_\tau \mid z_\tau) q_\theta(z^{\mathrm{f}}_\tau \mid z_\tau^{\mathrm{s}}) q_\theta(z_\tau^{\mathrm{s}} \mid \pi)$ in our proposed framework.
Note that, given the distribution $q_\theta(z_\tau^{\mathrm{s}} \mid \pi)$ over the slow states, all distributions required to compute the EFE $\mathcal{G}(\tau)$ can be obtained using the world model, and thus $\mathcal{G}(\tau)$ becomes computable.
Here, we replace the policy $\pi$ with an abstract action $\hat{A} \in \{\hat{A}_{n}\}_{n=1}^{K^{N_q}}$, and express the distribution $q_\theta(z_\tau^{\mathrm{s}} \mid \pi)$ as follows:
\begin{equation}
    \label{eq:approx-slow}
    \begin{aligned}
        q_\theta(z_\tau^{\mathrm{s}} \mid \pi)&\approx q_\theta(s_\tau^{\mathrm{s}} \mid d^\mathrm{s}_\tau, \hat{A}) q_\psi(d^\mathrm{s}_\tau\mid \hat{A}).\\
    \end{aligned}
\end{equation}
In this way, we can use the abstract world model $\mathcal{W}_\psi$ to predict the slow deterministic state $d^\mathrm{s}_\tau$ at $\tau = t + h$ from the abstract action $\hat{A}$.
Using the predicted deterministic state $d^\mathrm{s}_\tau$, we can obtain the slow prior $q_\theta(z_\tau^{\mathrm{s}} \mid \pi)$ and compute the EFE.
When computing the EFE, the prior preference $p(o_{\tau} \mid o_{\text{pref}})$ is assumed to follow a Gaussian distribution $\mathcal{N}(o_{\text{pref}}, \sigma^2)$ with mean ${o_{\text{pref}}}$ and variance $\sigma^2$. Therefore, the EFE can be written as follows:
\begin{equation}
    \label{eq:approx-efe}
    \begin{aligned}
        \mathcal{G}(\tau)\approx&-\mathbb{E}_{q_\theta(o_\tau, z_\tau \mid \pi)}[\log q_\theta(s^{\mathrm{f}}_\tau \mid z_\tau^\mathrm{s}, o_\tau )-\log q_\theta(s^{\mathrm{f}}_\tau \mid z_\tau^\mathrm{s})]\\
        &-\mathbb{E}_{q_\theta(o_\tau, z_\tau \mid \pi)}[{- \gamma{(o_{\tau}- o_{\text{pref}}})^2}],
    \end{aligned}
\end{equation}
where $\gamma = 1/2\sigma^2$ is the preference precision, which balances the epistemic and extrinsic values, and the expectations in the EFE are approximated via Monte Carlo sampling \cite{igari2024selection}.

To generate actual robot actions, we first use the abstract world model $\mathcal{W}_\psi$ to predict the slow deterministic states $\{d^\mathrm{s}_{t+h,n}\}_{n=1}^{K^{N_q}}$ at $h$ steps into the future for all abstract actions $\{\hat{A}_{n}\}_{n=1}^{K^{N_q}}$, given the current world model state $z_t$. 
Next, we predict slow hidden states $\{z^\mathrm{s}_{t+h,n}\}_{n=1}^{K^{N_q}}$ based on the predicted slow deterministic states by using \eqref{eq:approx-slow}.
Then, for each predicted state, we compute the EFE and select the abstract action that yields the minimum EFE. 
The selected abstract action is then decoded into an action sequence $\hat{a}_{t:t+h}$ by the action model, and the robot executes this sequence.

\section{EXPERIMENTS}

\subsection{Environment Setup}

\begin{figure}[t]
    \centering
    \includegraphics[width=0.92\linewidth]{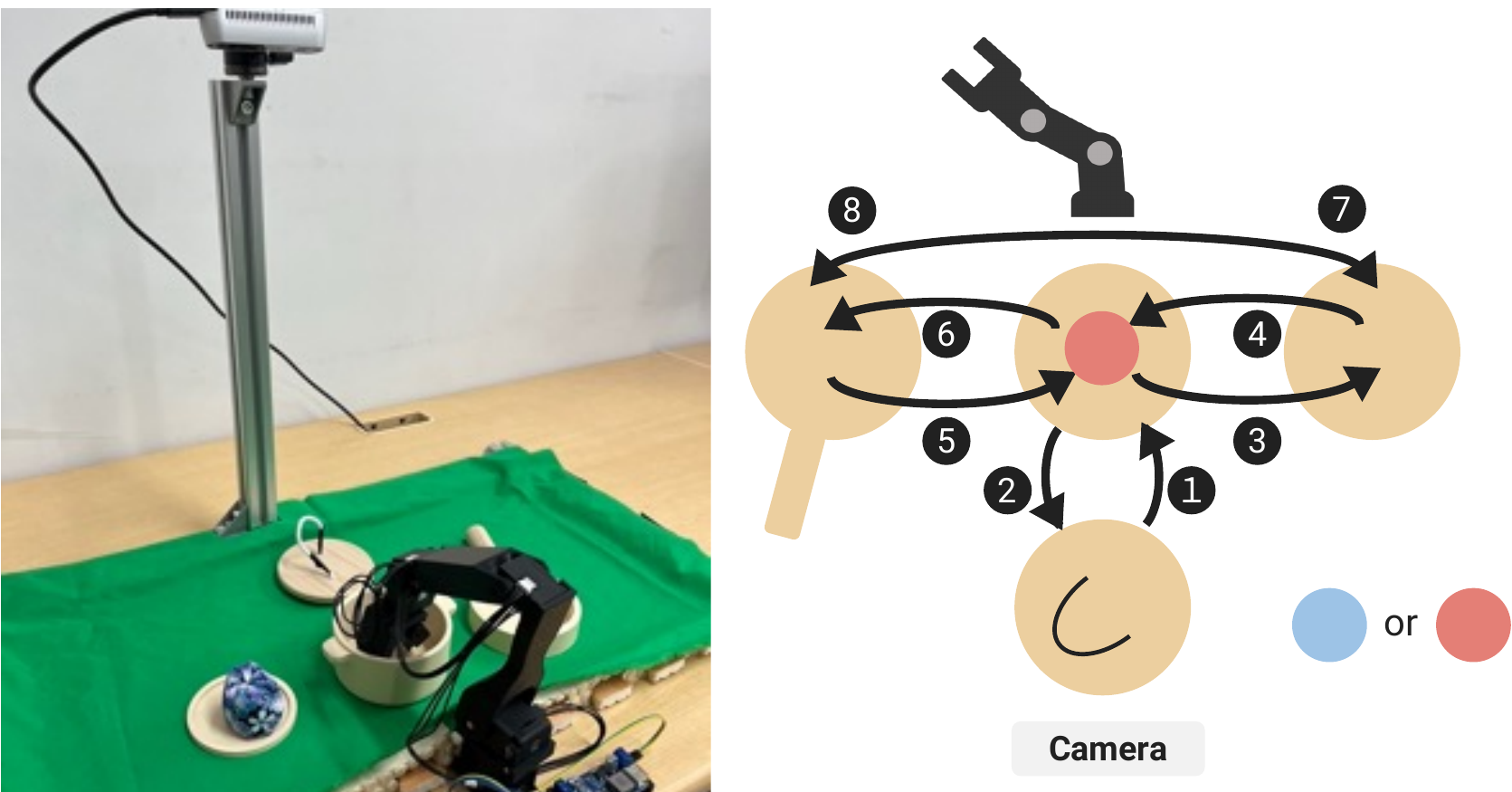}
    \caption{Experimental environment (left) and policy patterns included in the collected dataset (right).
    The environment contains either a blue ball, a red ball, or both.
    The dataset includes demonstrations of eight different policy patterns involving the movement of the lid and the balls.}
    \label{fig:environment}
\end{figure}

To investigate whether the proposed framework enables both goal achievement and exploration in real-world environments—where multiple objects can be manipulated and uncertainty arises from their placement—we conducted an experiment using a robot shown in Fig. \ref{fig:environment} (left) \cite{koch2024lowcost, cadene2024lerobot}.
The robot had six degrees of freedom, one of which is the gripper.
A camera (RealSense Depth Camera D435; Intel) was mounted opposite to the robot to capture a view of both the robot and its environment.
From the viewpoint of the camera, a simple dish, a pot, and a pan were placed on the right, center, and left, respectively, and a pot lid was placed closer to the camera than the center pot.
Additionally, the environment was configured such that a blue ball, a red ball, or both could be present.
Note that, therefore, uncertainty arose when the lid was closed, as the pot might or might not contain a blue or red ball in this environment.

As training data, we collected object manipulation data by demonstrating the predetermined eight patterns of policies (Fig. \ref{fig:environment}(right)). 
Each demonstration consists of a sequence of two patterns of policies. For all valid combinations—excluding those in which the policy would result in no movement (e.g., performing action 3 twice in a row)—we collected five demonstrations per combination by teleoperating the robot in a leader–follower manner.
There are $36$ valid action combinations for environments containing either a blue ball or a red ball, and $72$ combinations for environments containing both.
Each sequence contains $100$ time steps of joint angles and camera images recorded at $5$ Hz.  
Therefore, each pattern of policies had roughly $50$ time steps.
The original RGB images were captured, resized and clipped to $64 \times 80$. In this experiment, the robot action $a_t$ is defined as the absolute joint angle positions, and the observation $o_t$ is defined as the camera image.

\subsection{Interpretation of the Model Components}

In this experiment, we expected the slow hidden states $z_t^\mathrm{s}$ to represent the overarching progress of the task, such as where the balls and the lid were placed. 
In contrast, we expected the fast hidden states $z_t^\mathrm{f}$ represents more immediate, transient information.
On the other hand, we expected abstract actions $A_t$ to represent a meaningful behavior learned from the demonstration data. In an ideal case, an abstract action corresponds to one of the eight policy patterns in Fig. \ref{fig:environment}(right), such as moving the ball from the dish to the pan.

\subsection{Experimental Criteria}
\textbf{Capability of abstract world model:}
We evaluated the capability of the abstract world model.
First, we compared the computation time of our proposed framework against that of conventional deep active inference approaches \cite{Mazzaglia2021,sajid2021exploration,igari2024selection}, which predicts future states with the world model by sequentially inputting the action sequence $\hat{a}_{0:h}$ reconstructed from an abstract action $\hat{A}$ via the action model.

Second, we evaluated whether different predictions can be generated from the same initial state for each abstract action learned by the action model. We also examined whether the observed outcomes resulting from executing actual actions generated from a specific abstract action  are consistent with the predictions made by the abstract world model.

\textbf{Goal achievement performance:}
 
We evaluated the success rate on ball- (140 trials) and lid-manipulation (24 trials) tasks with varying object configurations, such as moving a particular ball or manipulating a lid. 
A trial was considered successful if the target object was placed in its specified goal position within $50$ time steps.

\textbf{Environment exploration:}
We evaluated whether the proposed framework can generate not only goal-directed actions but also exploratory actions from an uncertain initial situation.
To this end, we set up a scenario in which the blue ball is initially placed in the pan and the lid is closed, creating uncertainty about whether the red ball is present inside the pot.
In this scenario, when taking an exploratory action, it was expected that the robot would open the lid to resolve the uncertainty.

\subsection{Baseline and Ablation}
In the goal-achievement performance experiment, we compared our proposed framework with a baseline and two ablations described as follows:
\begin{itemize}
    \item \textbf{Goal-conditioned diffusion policy (GC-DP). } As a baseline, we implemented a diffusion policy with a U-Net backbone \cite{ronneberger2015u,chi2023diffusion}. In our implementation, this policy predicted a $48$-step future actions based on the two most recent observations and a goal observation. To stabilize actions, we apply an exponential moving average of weight $0.7$ to the generated actions.
    \item \textbf{Non-hierarchical. As an ablation study, the world model is replaced by a non-hierarchical dynamics model \cite{Hafner2020}. In this variant, the hidden state $z_t$ consists of a single-level deterministic state $d_t$ and a stochastic state $s_t$, where the deterministic state is computed using a gated recurrent unit \cite{chung2014empirical}.} 
    \item \textbf{No abstract world model (AWM).} As an ablation study, the robot does not use the abstract world model for planning. Instead, it calculates the EFE directly over actual action sequences decoded by the action model.
\end{itemize}
We did not perform an ablation on the action model itself, as our framework relies on it to generate the set of candidate actions (either abstract or actual) for evaluation, making it a core, indispensable component.

\section{RESULTS}
\subsection{Capability of abstract world model}
\begin{figure}[t]
    \centering
    \includegraphics[width=0.93\linewidth]{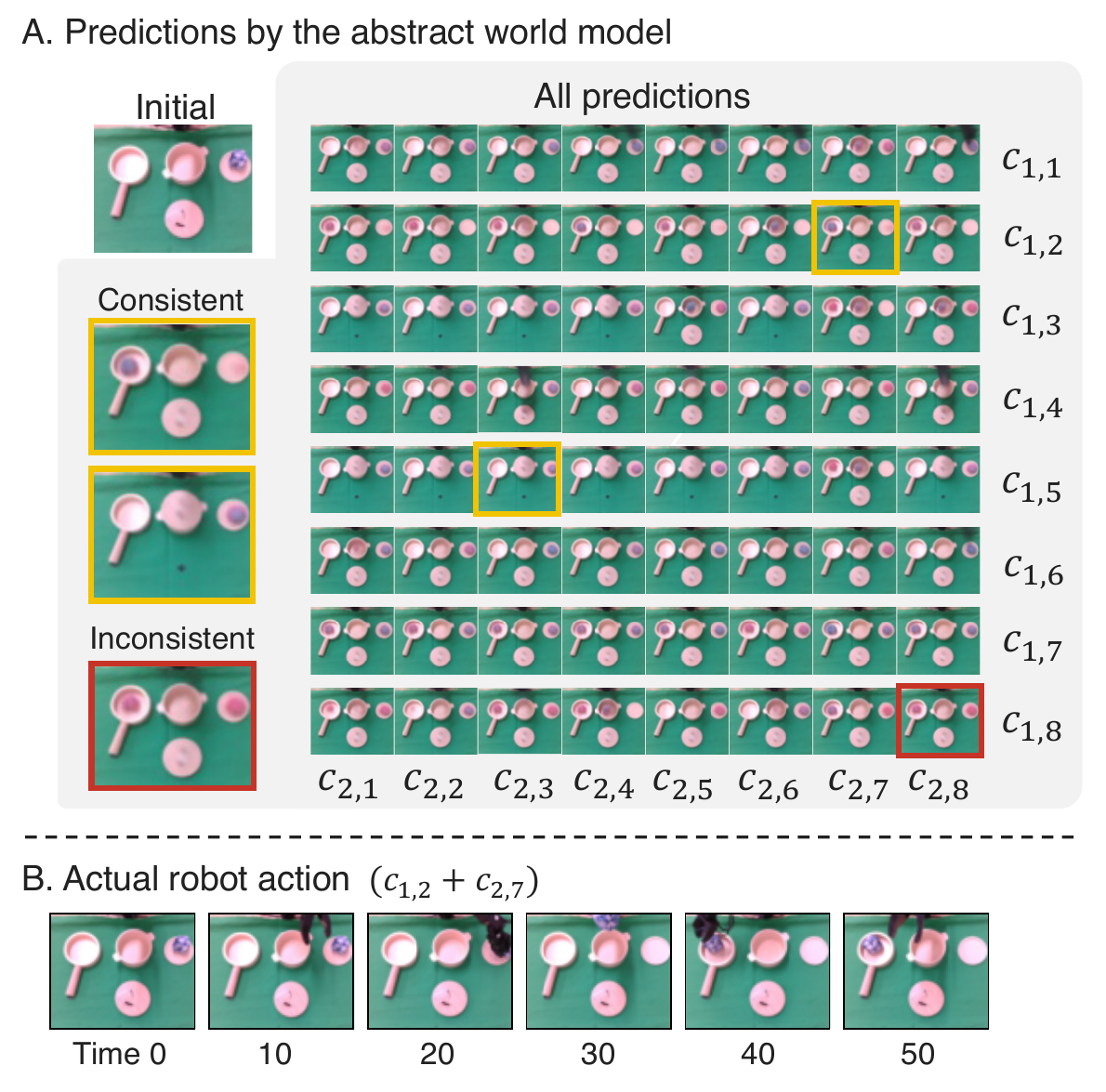}
    \caption{Example of predicted observations using the abstract world model and actual robot actions.
    (A) Predicted observations for each abstract action.
     Here, each $c_{i,j}$ denotes the $j$‑th code in the $i$‑th layer of the action model.
    The yellow box highlights an example prediction that is consistent with the initial observation, while the red box indicates an inconsistent prediction.
    (B) Actual observations corresponding to the action sequence generated from the abstract action $\hat{A}$ represented by $\hat{A}=c_{1,2} + c_{2,7}$ at each time step.
    }
    \label{fig:predicted-obs}
\end{figure}

Our proposed framework required only $2.37 \text{ ms}$ to evaluate all candidate abstract actions, in contrast to $71.8\text{ ms}$ for a sequential evaluation of conventional deep active inference approaches. This demonstrates the higher computational tractability of our proposed framework.

As shown in Fig. \ref{fig:predicted-obs}, different abstract actions lead to distinct predictions.
Moreover, for example, by using an action sequence generated from the abstract action represented by $c_{1,2} + c_{2,7}$, the ball was successfully moved from the dish to the pan, consistent with the predicted observation (Fig. \ref{fig:predicted-obs}).
These results suggest that the abstract world model has learned the dependency between abstract actions and the resulting state transitions, even without directly referring to actual action sequences. However, the prediction associated with the abstract action $\hat{A}$ represented by $\hat{A}=c_{1,8} + c_{2,8}$ in Fig. \ref{fig:predicted-obs} shows red balls placed on both the dish and the pan, which is inconsistent with the initial condition in which only a blue ball was present. This abstract action corresponded to moving a ball from the center pot to the pan. Since this action was not demonstrated when the pot was empty, the abstract world model may have learned incorrect dependencies for unlearned action–environment combinations.

\subsection{Goal achievement performance}
\begin{table}[t]
\centering
\caption{Success rate (\%).}
\begin{tabular}{l|cc|cc|c}
\hline
\multirow{2}{*}{Manipulation target} & \multicolumn{2}{|c|}{Ball} & \multicolumn{2}{|c|}{Lid} & \multirow{2}{*}{Total}\\
\cline{2-5} 
 & Red & Blue & Opening & Closing & \\
\hline
\hline
\textbf{Proposed} & $\boldsymbol{61.4}$ & $\boldsymbol{74.3}$ & $75.0$ & $\boldsymbol{100.0}$ & $\boldsymbol{70.7}$\\
GC-DP & $18.6$ & $25.7$ & $25.0$ & $50.0$ & $24.4$\\
Non-hierarchical & $41.4$ & $51.4$ & $75.0$ & $58.3$ & $51.2$\\
No AWM & $40.0$ & $21.4$ & $\boldsymbol{83.3}$ & $66.7$ & $37.2$\\
\hline
\end{tabular}
\label{tab:success-rate}
\end{table}
\begin{figure}[t]
    \centering
    \includegraphics[width=0.85\linewidth]{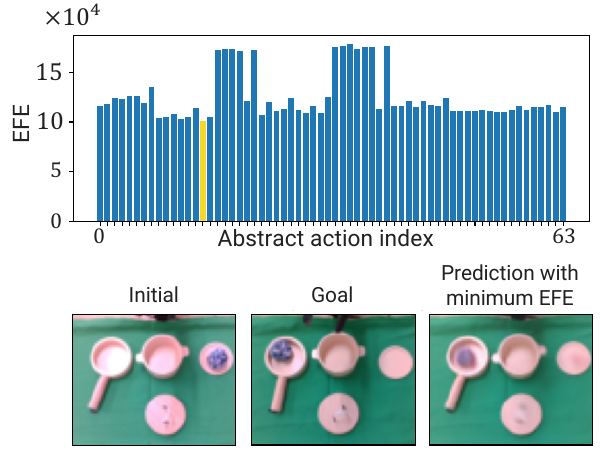}
    \caption{
    Example of EFE computed for each abstract action.
    \textbf{Top}: EFE values computed for all 64 abstract actions. The action with the lowest EFE is highlighted as a yellow bar.
    \textbf{Bottom}: From left to right, the images show the initial observation, goal, and predicted observation resulting from the abstract action with the lowest EFE.
    }
    \label{fig:efe}
\end{figure}

Table \ref{tab:success-rate} shows the success rates of our proposed framework on goal-directed action generation, evaluated on tasks involving specific ball and lid manipulations. The proposed method outperformed the baseline and the ablations across all goal conditions except the Lid-Opening goal, achieving a total success rate of over 70\%. As a qualitative example, Fig. \ref{fig:efe} illustrates the EFE calculation for a scenario where the goal is to move a ball from a dish to a pan. The abstract action with the lowest EFE correctly predicts the desired outcome, and executing the actual actions derived from this abstract action led to successful task completion. This overall result confirms that selecting abstract actions by minimizing the EFE is effective for goal achievement.

The failures in our framework were mainly due to inconsistent world model predictions, which misled the robot into believing an inappropriate action would succeed. For example, the proposed framework selected actions to grasp nothing but place the (non-grasped) target object at the appropriate location. In contrast, the GC-DP, Non-hierarchical, and No AWM all exhibited lower success rates. The GC-DP frequently failed in grasping and placing objects. Both ablations suffered from more prediction inconsistencies than our full model, highlighting the importance of temporal hierarchy and action/state abstraction. The lower performance of the No AWM ablation suggests that action abstraction was a particularly critical component for success.

\subsection{Environment exploration}

\begin{table}[t]
\setlength{\tabcolsep}{8pt}
\centering
\caption{EFE values for two representative abstract actions (goal-directed and exploratory) in the uncertain scenario.}
\begin{tabular}{l|c|c}
\hline
{Preference precision}  & Goal-directed  & Exploratory\\
\hline
\hline
$\gamma=10^2$&  $\boldsymbol{4.21 \times 10^4}$ & $14.5\times10^4$ \\
\hline
$\gamma=10^{-4}$ & $-4.67 \times 10^0$  & $\boldsymbol{-6.11 \times 10^0}$  \\
\hline
\end{tabular}
\label{tab:goal-explore-efe}
\end{table}

For simplicity, we computed the EFE for two abstract actions: moving the blue ball from the pan to the dish (goal-directed), and opening the lid (exploratory), as summarized in Table \ref{tab:goal-explore-efe}.
When preference precision $\gamma$ was set to $10^2$, the EFE for the goal-directed action became lower, and thus the robot moved the blue ball from  the pan to the dish.
In contrast, when preference precision $\gamma$ was set to $10^{-4}$, the EFE for the exploratory action became lower, and thus the robot opened the lid. 
These results indicate that the proposed framework can assign high epistemic value to exploratory actions that provide new information, and that exploratory actions can be induced by appropriately adjusting the preference precision $\gamma$.

\section{CONCLUSIONS}
\label{sec:conclusion}
In this work, we introduced a deep active‑inference framework that combines a temporally‑hierarchical world model, an action model utilizing vector quantization, and an abstract world model.
By capturing dynamics in a temporal hierarchy and encoding action sequences as abstract actions, the framework makes the action selection based on active inference computationally tractable.
 Real-world experiments on object-manipulation tasks demonstrated that the proposed framework outperformed the baseline in various goal-directed settings, as well as the ability to switch from goal-directed to exploratory actions in uncertain environments.

Despite these promising results, several challenges remain: 1) The action model used a fixed sequence length, which may not be optimal. 2) The model's predictive capability decreases for action-environment combinations not present in the dataset. 3) While we validated the capability to take exploratory actions, we did not evaluate their effectiveness in solving tasks and the switching to exploratory behavior still relies on a manually tuned hyperparameter.

Future work will focus on extending the framework to address these limitations. An immediate step is to evaluate our framework in environments that require multi-step action selection and where exploration is necessary to solve the task. Other promising directions include developing a mechanism for adaptive switching between goal-directed and exploratory modes, and extending the action model to represent variable-length action sequences. Ultimately, this work represents a significant step toward the long-term goal of creating more capable robots that can operate effectively in uncertain real-world environments such as household tasks by leveraging both goal-directed and exploratory behaviors.



\appendices

\section{EFE DERIVATION}

\label{appendix:efe}
We show the detailed derivation of EFE in our framework:
\begin{equation}
    \begin{aligned}
        \mathcal{G}(\tau)
        =&-\mathbb{E}_{q_\theta(o_\tau, z_\tau \mid \pi)}[\log q_\theta(z_\tau \mid o_\tau, \pi)-\log q_\theta(\mathrm{z_\tau} \mid \pi)]\\
        &-\mathbb{E}_{q_\theta(o_\tau, z_\tau \mid \pi)}[\log p(o_\tau \mid  o_{\text{pref}})]\\
        =&-\mathbb{E}_{q_\theta(o_\tau, z_\tau \mid \pi)}[\log q_\theta(z_\tau^{\mathrm{f}} \mid z^{\mathrm{s}}_\tau , o_\tau )q_\theta(z^{\mathrm{s}}_\tau\mid  \pi)\\
        &-\log q_\theta(z_\tau^{\mathrm{f}} \mid z^{\mathrm{s}}_\tau )q_\theta(z^{\mathrm{s}}_\tau\mid  \pi)]\\
        &-\mathbb{E}_{q_\theta(o_\tau, z_\tau \mid \pi)}[\log p(o_\tau \mid  o_{\text{pref}})]\\
        =&-\mathbb{E}_{q_\theta(o_\tau, z_\tau \mid \pi)}[\log q_\theta(z_\tau^{\mathrm{f}} \mid z^{\mathrm{s}}_\tau, o_\tau )-\log q_\theta(z_\tau^{\mathrm{f}} \mid z^{\mathrm{s}}_\tau)]\\
        &-\mathbb{E}_{q_\theta(o_\tau, z_\tau \mid \pi)}[\log p(o_\tau \mid  o_{\text{pref}})]\\
        =&-\mathbb{E}_{q_\theta(o_\tau, z_\tau \mid \pi)}[\log q_\theta(s^{\mathrm{f}}_\tau \mid d^\mathrm{f}_\tau, o_\tau )q_\theta(d^\mathrm{f}_\tau\mid z_\tau^{\mathrm{s}})\\
        &-\log q_\theta(s^{\mathrm{f}}_\tau \mid d^\mathrm{f}_\tau )q_\theta(d^\mathrm{f}_\tau\mid z_\tau^{\mathrm{s}})]\\
        &-\mathbb{E}_{q_\theta(o_\tau, z_\tau \mid \pi)}[\log p(o_\tau \mid  o_{\text{pref}})]\\
        \approx&-\mathbb{E}_{q_\theta(o_\tau, z_\tau \mid \pi)}[\log q_\theta(s^{\mathrm{f}}_\tau \mid z_\tau^\mathrm{s}, o_\tau )-\log q_\theta(s^{\mathrm{f}}_\tau \mid z_\tau^\mathrm{s})]\\
        &-\mathbb{E}_{q_\theta(o_\tau, z_\tau \mid \pi)}[\log p(o_\tau \mid  o_{\text{pref}})].
    \end{aligned}
    \label{eq:efe-detailed}
\end{equation}

\section{ADDITIONAL EXPERIMENTS}
To validate the scalability of our framework, we further evaluated our framework on the CALVIN D benchmark \cite{mees2022calvin}, which provides various unstructured human data.
Although this environment can serve language goal conditioning, we used only image-based goal conditioning.  

For this environment, we compared our proposed framework with the GC-DP. The evaluation was conducted on eight tasks: move\_slider\_left/right (Slider), open/close\_drawer (Drawer), turn\_on/off\_lightbulb (Lightbulb), and turn\_on/off\_led (LED). A trial was considered successful if the task was completed within $150$ timesteps. Our proposed framework used the same hyperparameters as in our primary experiments, but the GC-DP was trained to predict a $28$-step future action sequence from a four-step observation history and re-planned every $16$ steps.

As shown in Table \ref{tab:calvin-results}, our proposed method consistently outperformed GC-DP on the Slider and Drawer tasks, as well as on the average success rate across all tasks. 
These results suggest that our approach, which leverages a temporally hierarchical world model and abstract actions, is robust and effective not only in our primary setup but also in more complex, long-horizon manipulation scenarios.
\begin{table}[t]
\vspace{-3mm} 
\centering
\caption{Success rate in CALVIN Environmet (\%).}
\begin{tabular}{l|cccc|c}
\hline
task  & Slider & Drawer & Lightbulb & LED &  Total\\
\hline
\hline
\textbf{Proposed} & $\boldsymbol{43.8}$ & $\boldsymbol{93.8}$ & $0.0$ & $11.8$ & $\boldsymbol{37.5}$\\
GC-DP & $1.6$ & $68.3$ & \boldsymbol{$52.6$} & \boldsymbol{$16.7$} & $34.8$\\
\hline
\end{tabular}
\label{tab:calvin-results}
\end{table}

\section{HYPER PARAMETERS}
We show hyperparameters in our experiments in Table \ref{tab:hparams}.
\label{sec:hyper-param}
\begin{table}[t]
\centering
\caption{Hyperparameters of our proposed framework}
\begin{tabular}{lcc}
\hline
 \textbf{Name} & \textbf{Symbol} & \textbf{Value} \\
\hline
World Model \\
\hline
 Training data sequence length & --- & 75 \\
Slow dynamics   & & \\
 \quad Deterministic state dimensions& --- & 32 \\
\quad Stocahstic state dimensions $\times$ classes & --- & $4\times 4$ \\
 \quad Time constant & --- & 32 \\
Fast dynamics   &  & \\
 \quad Deterministic state dimensions& --- & 128 \\
\quad Stocahstic state dimensions $\times$ classes & --- & $8\times 8$ \\
 \quad Time constant & --- & 4 \\
 KL balancing & $w$ & 0.8 \\
\hline
Action Model \\
\hline
 Layers of MLP & --- & 2 \\
 Hidden dimensions of MLP & --- & 128 \\
 Action sequence length & $h$ & 50\\
 Codebook size & $K$ & 8 \\
 Abstract action dimensions & --- & 32 \\
 Learning coefficients & $\lambda_\text{MSE}, \lambda_\text{commit}$ & $1.0, 5.0$ \\
\hline
Abstract World Model \\
\hline
 Layers of MLP & --- & 2 \\
 Hidden dimensions of MLP & --- & 512 \\
\hline
\end{tabular}
\label{tab:hparams}
\end{table}


\bibliographystyle{IEEEtran}

\bibliography{IEEEabrv,reference}

\addtolength{\textheight}{-12cm}   

\end{document}